\documentclass{article}

\usepackage{PRIMEarxiv}

\usepackage[utf8]{inputenc} 
\usepackage[T1]{fontenc}    
\usepackage{hyperref}       
\usepackage{url}            
\usepackage{booktabs}       
\usepackage{float}
\usepackage{amsfonts}       
\usepackage{amsmath}
\usepackage{amssymb}
\usepackage{nicefrac}       
\usepackage{microtype}      
\usepackage{lipsum}
\usepackage{todonotes}
\usepackage{fancyhdr}       
\usepackage{graphicx}       
\usepackage{subcaption}
\graphicspath{{media/}}     
\usepackage{orcidlink}

\usepackage[pagewise]{lineno}
\usepackage{todonotes}
\nolinenumbers

\usepackage{amsthm}

\DeclareMathOperator*{\argmin}{arg\,min}

\title{Picard-KKT-hPINN: Enforcing Nonlinear Enthalpy Balances for Physically Consistent Neural Networks
}

\author{
  \orcidlink{0009-0005-3475-9351} Giacomo Lastrucci\\
  Process Intelligence Research Group\\
  Department of Chemical Engineering\\
  Delft University of Technology\\
   \And
  Tanuj Karia \\
  Process Intelligence Research Group\\
  Department of Chemical Engineering\\
  Delft University of Technology \\
   \And
  Zoë Gromotka \\
  Mathematical Physics Group\\
  Delft Institute of Applied Mathematics\\
  Delft University of Technology \\
   \And
  \orcidlink{0000-0001-8885-6847} Artur M. Schweidtmann\thanks{corresponding author}\\
  Process Intelligence Research Group\\
  Department of Chemical Engineering\\
  Delft University of Technology \\
  \texttt{a.schweidtmann@tudelft.nl} \\
}

\begin{document}
\maketitle

\begin{abstract}
Neural networks (NNs) are widely used as surrogate models but they do not guarantee physically consistent predictions thereby preventing adoption in various applications. We propose a method that can enforce NNs to satisfy physical laws that are nonlinear in nature such as enthalpy balances. Our approach, inspired by Picard’s successive approximations method, aims to enforce multiplicatively separable constraints by sequentially \textit{freezing} and projecting a set of the participating variables. We demonstrate our Picard-KKT-hPINN for surrogate modeling of a catalytic packed bed reactor for methanol synthesis. Our results show that the method efficiently enforces nonlinear enthalpy and linear atomic balances at machine-level precision. Additionally, we show that enforcing conservation laws can improve accuracy in data-scarce conditions compared to \textit{vanilla} multilayer perceptron.
\end{abstract}


\section{Introduction}
Surrogate modeling plays a crucial role in simplifying and approximating complex physical models, making them suitable for large-scale simulations and optimization studies of industrial relevance.  Machine learning models, such as neural networks (NNs), are particularly well-suited for this purpose due to their simplicity and strong regression capabilities~\cite{McBride2019_OverviewSurrogateModelinga}. However, despite exceptional advancements in machine learning, issues and skepticism regarding the black-box nature and physical inconsistency of these models hinder the adoption of machine learning-based tools (and, more broadly, artificial intelligence) in industrial applications~\cite{Bedue2021_Canwetrust, Schweidtmann2021_Obeyvaliditylimits}.\\
To mitigate this limitation, significant research has been carried out to enforce known mechanistic relationships between inputs and predictions in NNs. \textit{Soft-constrained} neural networks represent an approach in which physical equations are included as penalty terms in the loss function~\cite{Raissi2019_Physicsinformedneural,Wu2023_ApplicationPhysicsInformed}. This method does not require any modification of the neural network architecture, but it does not guarantee the exact satisfaction of the constraints. When exact adherence to constraints such as closure of balances is required, \textit{hard-constrained} neural networks become relevant. Models belonging to this class aim to strictly satisfy given equations or symmetries underlying the original system (constrained learning).\\
Different methodologies have been proposed for constrained learning. For instance, Beucler et al. (2019) and Donti et al. (2019) apply a method known as \textit{prediction and completion}, where an NN is trained to predict a subset of variables, while the remaining ones are determined by solving a (non)linear system during training~\cite{Beucler2019_EnforcingAnalyticConstraints, Donti2021_DC3learningmethod}. Researchers in optimization proposed to learn solutions to constrained optimization problems leveraging Lagrangian duality~\cite{Fioretto2021_LagrangianDualityConstrained}; however, aspects of this approach still rely on soft constraints. Chen et al. (2021) introduced a projection step to guide the learning process by enforcing adherence to discretized differential equations~\cite{Chen2021_Theoryguidedhard}. Based on a similar concept, Chen, et al. (2024) presented KKT-hPINN to adjust neural network predictions, ensuring compliance with linear algebraic constraints~\cite{Chen2024_PhysicsInformedNeural}. Recently, Mukherjee and Bhattacharyya (2024) proposed to train constrained neural network using an NLP solver such as IPOPT~\cite{Mukherjee2024_developmentsteadystate}. Likewise, Schweidtmann et al. (2019) have used deterministic global optimization to compute guaranteed worst-case accuracies of NNs~\cite{Schweidtmann2019_DeterministicGlobalProcess}. However, the use of local or global NLP solvers can be memory and time-intensive for NNs that have a large number of parameters and training data points. Overall, current approaches (1) rely on external solvers that increase the computational cost to train and evaluate the NN or (2) are limited to enforcing linear equality constraints.\\
To address these challenges, we introduce a computationally efficient method that enforces relevant physical laws in the form of nonlinear algebraic equations. We extend the KKT-hPINN framework by considering local projections and numerical techniques to ensure exact constraint satisfaction at machine-level precision. Drawing inspiration from Picard’s successive approximation method for solving nonlinear PDEs, we name our model \textit{Picard-KKT-hPINN}.

\section{Methods}
We introduce Picard-KKT-hPINN, a method to enforce algebraic nonlinear constraints, such as energy balances, in NN predictions. It extends the KKT-hPINN approach with local projections and numerical techniques such as variable-freezing, while maintaining computational efficiency.

\subsection{Problem statement}
Given a dataset $(\mathbf{x}_i, \mathbf{y}_i)_{i=1, \dots, N}$, representing the input-output behavior of a physical system (e.g., a unit operation), our goal is to train a neural network (NN) to approximate the underlying model governing this system. In some cases, algebraic equations describing some relationship between input and output are known beforehand. We aim to develop an NN $f_{\theta}$ such that the predicted output $\hat{\mathbf{y}} = f_{\theta}(\mathbf{x})$ satisfies a set of algebraic constraints $c_k(\mathbf{x}, \mathbf{y}) = 0, \, k = 1, \dots, N_C$, which may represent conservation laws or design specifications. The number of constraints ($N_C$) must not exceed the dimensionality of the neural network's output ($N_O$). This ensures the system is underdetermined ($N_C < N_O$), allowing the network to learn from the data while adhering to these constraints.

\subsection{Background: KKT-hPINN}
Chen et al. (2024) proposed a model denoted as KKT-hPINN to enforce linear relationships in NNs~\cite{Chen2024_PhysicsInformedNeural}. In KKT-hPINN, the (physically inconsistent) prediction of a generic multilayer perceptron is corrected by an orthogonal projection onto the linear feasible region described by a system of linear equations and geometrically defining a hyperplane. 
Specifically, the prediction $\hat{\mathbf{y}}$ is corrected so as to respect a linear system by solving the following optimization problem:

\begin{equation}
\label{eq:quadr_progr}
\begin{aligned}
\Tilde{\mathbf{y}} =& \argmin_{\mathbf{y}} \frac{1}{2} \| \mathbf{y} - \hat{\mathbf{y}} \|^2 \\
\text{s.t.} \quad & \mathbf{A} \mathbf{x} + \mathbf{B} \mathbf{y} = \mathbf{b}
\end{aligned}
\end{equation}

The corrected, physically consistent, solution $\Tilde{\mathbf{y}}$ is achieved by solving the Karush-Kuhn-Tucker (KKT) conditions of this optimization problem (Eq.~\ref{eq:quadr_progr}), thus the name KKT-hPINN:

\begin{equation}
    \label{eq:correction}
    \Tilde{\mathbf{y}} = \mathbf{A}^* \mathbf{x} + \mathbf{B}^* \Hat{\mathbf{y}} + \mathbf{b}^*
\end{equation}

Considering the identity matrix $\mathbf{I}$, we define:

\begin{equation}
    \label{eq:correction_matrices}
    \begin{aligned}
    \mathbf{A}^* &= -\mathbf{B} (\mathbf{B} \mathbf{B}^\top)^{-1} \mathbf{A} \\
    \mathbf{B}^* &= \mathbf{I} - \mathbf{B}^\top (\mathbf{B} \mathbf{B}^\top)^{-1} \mathbf{B} \\
    \mathbf{b}^* &= \mathbf{B}^\top (\mathbf{B} \mathbf{B}^\top)^{-1} \mathbf{b}
    \end{aligned}
\end{equation}

Hence, the projection matrices ($\mathbf{A}^*$,$\mathbf{B}^*$,$\mathbf{B}^*$) are computed analytically owing to the quadratic objective function and linear constraints. This projection acts as a physics-informed activation layer on the network output, providing a computationally efficient approach to enforce linear constraints. However, its applicability is limited to linear systems, making it unsuitable for scenarios with nonlinearities. For instance, while effective for modeling mass balances in chemical engineering, it struggles with enthalpy balances, which often involve nonlinear relationships.

\subsection{Local projection}
We generalize the linear constraints formulation in Eq.~\ref{eq:quadr_progr} to any nonlinear function in the input vector $\mathbf{x}$:

\begin{equation}
\label{eq:constraint}
c(\mathbf{x}, \mathbf{y}) = \mathbf{B} \mathbf{y} + g(\mathbf{x}) - \mathbf{b} = 0
\end{equation}

Since the input vector is known for every instance, we can consider it as a constant term and define $v(\mathbf{x})=\mathbf{b}-g(\mathbf{x})$. Moreover, the input vector does not participate as an optimization variable in the projection process and thus $g(\mathbf{x})$ can be any nonlinear. Hence, Eq.~\ref{eq:constraint} can be written as:

\begin{equation}
\label{eq:constraint_ref}
c(\mathbf{x}, \mathbf{y}) = \mathbf{B} \mathbf{y} - v(\mathbf{x}) = 0
\end{equation}

Where $v(\mathbf{x})$ is a vector-valued function $v: \mathbf{x} \subseteq \mathbb{R}^{N_I} \to \mathbb{R}^{N_C}$. In particular, $v$ is a nonlinear function of the $N_I$-dimensional input to the NN, meaning that its value is instance-dependent (an instance is a given datapoint $(\mathbf{x}_i, \mathbf{y}_i)$). Thus, the value of the constraint must be updated depending on each datapoint instance and cannot be defined offline. Additionally, we consider local (instance-specific) constraint definition by allowing the matrix $\mathbf{B}$ to depend on the output instance $\mathbf{y}_i$. To simplify the notation, we define $\mathbf{B}_i = \mathbf{B}(\mathbf{y}_i)$ and $\mathbf{v}_i = \mathbf{v}(\mathbf{x}_i)$ such that:

\begin{equation}
\label{eq:constraint_ref_i}
c(\mathbf{x}_i, \mathbf{y}_i) = \mathbf{B}_i \mathbf{y}_i - v_i = 0
\end{equation}

This formulation comprises a wider class of functions with respect to the linear case in Eq.~\ref{eq:quadr_progr} and is foundational for handling local nonlinearities. 
The computational cost to achieve local projection can be reduced by exploiting parallel computing. We perform batch training (with $BS$ being the number of elements in the batch) by building a rank-3 tensor $\mathbf{B} \in \mathbb{R}^{BS \times N_C \times N_O}$ holding a number $BS$ of local $\mathbf{B}_i$ matrices. Similarly, we construct a rank-2 tensor $\mathbf{V} \in \mathbb{R}^{BS \times N_C}$ to store the local $\mathbf{v}_i$ vectors. The local projection matrices are computed in parallel (Eq.~\ref{eq:B_V_star}), hence reducing the complexity of the most expensive computation, the matrix inversion, from $\mathcal{O}(BS \times N^3)$ to $\mathcal{O}(N^3)$.

\begin{equation}
\label{eq:B_V_star}
\begin{aligned}
\mathbf{B}^* &= \mathbf{I} - \mathbf{B}^T (\mathbf{B} \mathbf{B}^T)^{-1} \mathbf{B}, \\
\mathbf{V}^* &= \mathbf{B}^T (\mathbf{B} \mathbf{B}^T)^{-1} \mathbf{V}
\end{aligned}
\end{equation}

More specifically, considering the batch inversion of the tensor $\mathbf{B} \mathbf{B}^T$, the dimension $N$ corresponds to the number of constraints. Assuming tasks with $N_C = N < 10^3$, the inversion requires a negligible number of FLOPs on modern hardware. To the best of our knowledge, this dimensionality bound does not restrict the applicability of the method for the majority of surrogate modeling tasks in chemical engineering.

\subsection{Picard-KKT-hPINN}
Inspired by Picard's successive approximation method for nonlinear partial differential equations (PDEs), we develop a model to exactly enforce nonlinear constraints, including physical balances expressed as algebraic equations.

\begin{figure}[t]
    \centering
    \includegraphics[width=0.85\textwidth]{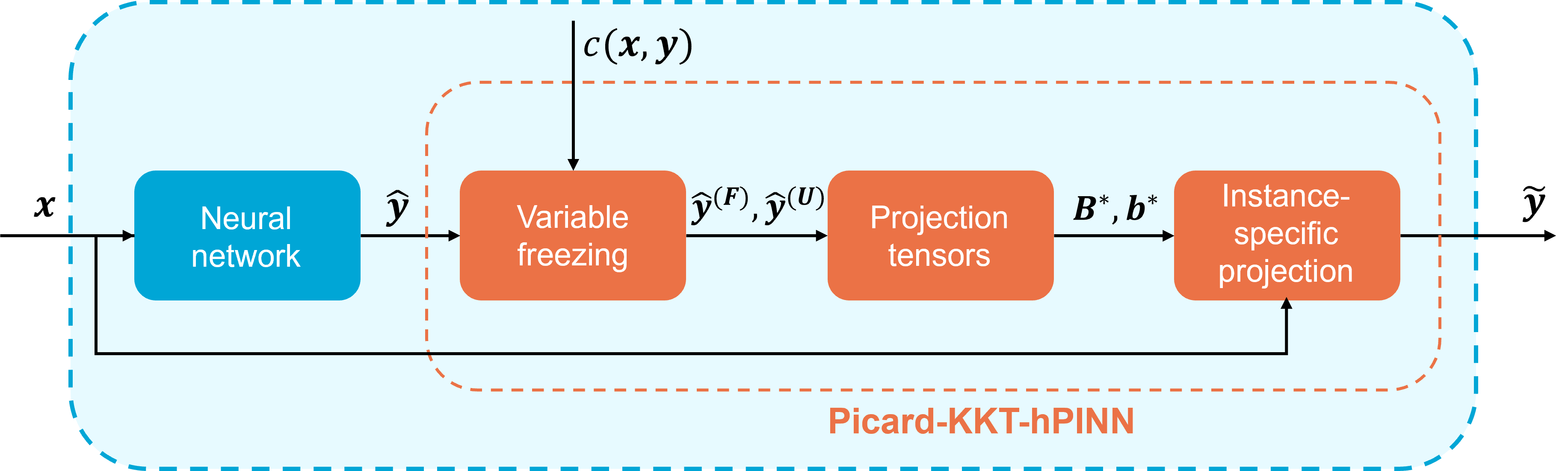}
    \caption{Picard-KKT-hPINN consists of a set of non-trainable layers that can be appended on every NN backbone (e.g., multilayer perceptrons, convolutional neural networks, transformers). Multiplicatively separable constraints (Eq.~\ref{eq:constraints_class}) are enforced exactly by partitioning the prediction vector (\textit{variable freezing}). Then, the prediction of the NN is projected onto the feasible space defined by the constraints \textit{c}. The layers composing Picard-KKT-hPINN are differentiable, hence, the NN is trained end-to-end and the architecture is equivalent at training and inference time.}
    \label{fig:Architecture}
\end{figure}

\subsubsection{Output partitioning and definition of the constraint class}
Given the NN's output vector $\mathbf{y} = [y_1, \dots, y_{N_O}] \in \mathbb{R}^{N_O}$, we partition its components as $\mathbf{y} = [\mathbf{y}^{(F)} \; \mathbf{y}^{(U)}]$. $\mathbf{y}^{(F)}$ is the subvector of \textit{frozen} variables, and $\mathbf{y}^{(U)}$ is the subvector of \textit{unfrozen} variables, with $\mathbf{y}^{(F)} = [y_{j_1}, \dots, y_{j_{N_F}}] \in \mathbb{R}^{N_F}$ and $\mathbf{y}^{(U)} = [y_{j_1}, \dots, y_{j_{N_U}}] \in \mathbb{R}^{N_O - N_F}$, where $N_U = N_O - N_F$.\\
We aim to enforce multiplicatively separable equality constraints of the form:

\begin{equation}
\label{eq:constraints_class}
c_k(\mathbf{x}, \mathbf{y}) = \mathbf{B}_k \mathbf{y} + F_k(\mathbf{y}^{(F)}) \cdot \mathbf{y}^{(U)} - v_k(\mathbf{x}) = 0, \quad k = 1, \dots, N_C
\end{equation}

The constraint includes a linear function in the output $\mathbf{y}$, a nonlinear function $v_k$ in the input $\mathbf{x}$, and a scalar multiplication between $F_k$, a nonlinear vector-valued function of the frozen variables $\mathbf{y}^{(F)}$, and the unfrozen variables $\mathbf{y}^{(U)}$. Note that $F_k: \mathbb{R}^{N_F} \rightarrow \mathbb{R}^{N_U}$.\\
Steady-state macroscopic conservation principles can generally be formulated as Eq.~\ref{eq:constraints_class}. Thus, the structure of the considered constraint is of particular interest for engineering and physics-based applications.\\
More generally, the constraints in Eq.~\ref{eq:constraints_class} can be expressed in matrix notation as:

\begin{equation}
\label{eq:constraints_class_matrix}
c(\mathbf{x}, \mathbf{y}) = \mathbf{B} \mathbf{y} + F(\mathbf{y}^{(F)}) \cdot \mathbf{y}^{(U)} - v(\mathbf{x}) = 0
\end{equation}

Here, $\mathbf{c}(\mathbf{x}, \mathbf{y})$ is the vector of $N_C$ constraints, and $\mathbf{F}(\mathbf{y}^{(F)})$ is an $N_C \times N_U$ matrix. We develop a method to ensure that the NN predictions exactly satisfy constraints belonging to the class described in Eq.~\ref{eq:constraints_class_matrix} by \textit{freezing} a partition of the output variables, resembling Picard’s iteration method.

\subsubsection{Variables freezing}
Given the output $\hat{\mathbf{y}}$ of an NN $f_{\theta}(\mathbf{x})$, we aim to guarantee the prediction to satisfy a constraint of the form as in Eq.~\ref{eq:constraints_class_matrix} by projecting it onto the feasible space. The projected prediction $\tilde{\mathbf{y}}$ can be computed as the solution of the following nonlinear program (NLP):

\begin{equation}
\label{eq:quad_program_freezing}
\begin{aligned}
\tilde{\mathbf{y}} =& \argmin_{\mathbf{y}} \|\mathbf{y} - \hat{\mathbf{y}}\|^2 \\
\text{s.t.} & \quad \mathbf{c}(\mathbf{x}, \mathbf{y}) = \mathbf{B}\mathbf{y} + \mathbf{F}(\mathbf{y}^{(F)}) \cdot \mathbf{y}^{(U)} - \mathbf{v}(\mathbf{x}) = 0
\end{aligned}
\end{equation}

Instead of relying on an external solver, we reformulate the NLP in Eq.~\ref{eq:quad_program_freezing} as a quadratic program (QP) solvable analytically. To achieve this, we assume the variables in $\mathbf{y}^{(F)}$ as \textit{correct} and thus set them constant. This variable-freezing allows us to reformulate Eq.~\ref{eq:constraints_class_matrix} as a linear constraint by defining $\mathbf{B}' = \mathbf{B} + \mathbf{F}(\mathbf{y}^{(F)})$:

\begin{equation}
\label{eq:reformulation}
\mathbf{c}(\mathbf{x}, \mathbf{y}) = \mathbf{B}'\mathbf{y} - \mathbf{v}(\mathbf{x}) = 0.
\end{equation}

Since Eq.~\ref{eq:reformulation} is the matrix form equivalent of Eq.~\ref{eq:constraint_ref_i}, the local projection is achieved as described in Section~2.3 and results in $\tilde{\mathbf{y}} = \mathbf{B}^* \hat{\mathbf{y}} + \mathbf{v}^*$, where the projection tensors can be computed in parallel during training.\\
The nonlinear term is handled by projecting only a partition of the output variables. We observe that the method is more efficient when $N_U < N_F$, meaning that more variables are projected to guarantee the nonlinear constraints. However, in general, the whole output vector can still potentially be projected, as linear terms may also be present. This numerical technique resembles Picard’s iteration method for solving nonlinear PDEs iteratively (hence the name Picard-KKT-hPINN), where some variables are kept constant from the previous iteration’s value to achieve a linear expression. Similarly, here we exploit the iterative training process to converge towards the ground truth data while guaranteeing that the prediction obeys the nonlinear constraint.

\section{Results and discussion}
\subsection{Case study and constraints}
Our proposed method is illustrated on a tubular packed-bed reactor for methanol synthesis, using the same first-principle modeling assumptions as in Lastrucci et al. (2024)~\cite{Lastrucci2024_Physicsinformedneural}. The surrogate models involve 10 inputs (inlet temperature, pressure, component flowrates, and coolant flowrate) and 10 outputs (outlet temperature, pressure, component flowrates, and hotspot temperature). The training and test datasets, consisting of 20,000 and 500 data points respectively, are generated by solving the mechanistic model under diverse conditions bounded by operational limits derived from flowsheet simulation.\\
We derive algebraic constraints underlying the steady state physical system that we aim to enforce in the surrogate model, such as atomic balance and total enthalpy balance. The system consists of four linear atomic balances (Eq.~\ref{eq:atomic_bal}) and one nonlinear equation for the enthalpy balance (Eq.~\ref{eq:energy_bal}):

\begin{equation}
\label{eq:atomic_bal}
\sum_{i=1}^{N_S} \dot{a}_{ij,\text{in}} - \sum_{i=1}^{N_S} \dot{a}_{ij,\text{out}} = 0, \quad j = \text{C, H, O, N}
\end{equation}

\begin{equation}
\label{eq:energy_bal}
\sum_{i=1}^{N_S} \dot{n}_{i,\text{in}} \tilde{h}_{i,\text{in}} - \sum_{i=1}^{N_S} \dot{n}_{i,\text{out}} \tilde{h}_{i,\text{out}} - \dot{Q} = 0
\end{equation}

In Eq.~\ref{eq:atomic_bal}, $\dot{a}_{ij,\text{in}}$ and $\dot{a}_{ij,\text{out}}$ represent the molar flowrates of atom $j$ carried by species $i$, entering and exiting the reactor, respectively, and $N_S$ is the number of chemical species involved (composed of four atoms, such as carbon, hydrogen, oxygen, and nitrogen). Being the atomic balances linear constraints, they can effectively be enforced using KKT-hPINN~\cite{Chen2024_PhysicsInformedNeural}. In Eq.~\ref{eq:energy_bal}, $\dot{n}_{i,\text{in}}$ and $\dot{n}_{i,\text{out}}$ represent the molar flowrates of species $i$, $h_{i,\text{in}}$ and $h_{i,\text{out}}$ are the molar-specific enthalpies, and $\dot{Q}$ is the rate of heat transfer out of the system. In general, the specific enthalpy $\tilde{h}_{i,\text{eval}} = \tilde{h}_{i,\text{eval}}(T_\text{eval})$ is a nonlinear function of the temperature, depending on the modeling choice for the specific heat at constant pressure $c_P$. \\
Since in the considered case study $\dot{Q} = \dot{n}_c \Delta\tilde{H}_\text{ev}(T_c)$ is a function of the input (coolant flowrate, $\dot{n}_c$)~\cite{Lastrucci2024_Physicsinformedneural}, the term $\sum_{i=1}^{N_S} \dot{n}_{i,\text{in}} \tilde{h}_{i,\text{in}} - \dot{Q}$ can be regarded as $v_k(\mathbf{x})$ in Eq.~\ref{eq:constraints_class}. More importantly, the term $-\sum_{i=1}^{N_S} \dot{n}_{i,\text{out}} \tilde{h}_{i,\text{out}}$ can be seen as the dot product in Eq.~\ref{eq:constraints_class}, with $F_k(\mathbf{y}^{(F)}) = \tilde{h}_{i,\text{out}} = \tilde{h}_i(T_\text{out})$ and $\mathbf{y}^{(U)} = \dot{n}_{i,\text{out}}$. Thus, our Picard-KKT-hPINN model can exactly enforce both the atomic (Eq.~\ref{eq:atomic_bal}) and the enthalpy balances (Eq.~\ref{eq:energy_bal}).

\subsection{Experiments and comparison}
\begin{figure}[t]
    \centering
    \begin{subfigure}[t]{0.55\textwidth}
        \centering
        \includegraphics[width=\textwidth]{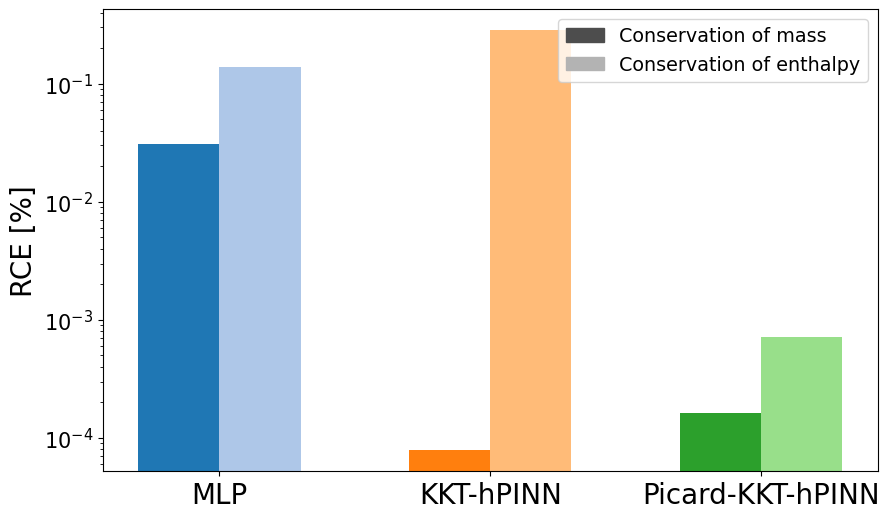}
        \caption{Relative conservation error (RCE)}
        \label{fig:RCE}
    \end{subfigure}
    \begin{subfigure}[t]{0.535\textwidth}
        \centering
        \includegraphics[width=\textwidth]{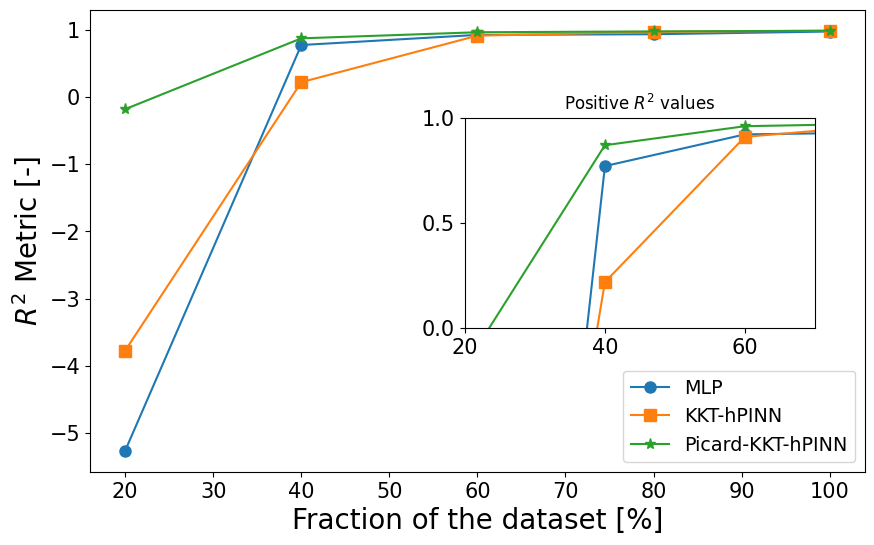}
        \caption{Data scarcity}
        \label{fig:data-scarcity}
    \end{subfigure}
    \caption{Comparison of relative conservation error (RCE) of mass and enthalpy balances (a) and regression performance in data scarcity conditions (b) over the three different models.}
    \label{fig:results}
\end{figure}
We train a shallow Picard-KKT-hPINN consisting of 1 hidden layer with 64 neurons and ReLU activation function, and we compare it against a KKT-hPINN and a multilayer perceptron (MLP) with the same backbone architecture. Every model is trained for 50,000 epochs using the Adam optimizer with a learning rate of $10^{-5}$ and a batch size of 2,000 samples. For reproducibility, the fully connected layers are deterministically optimized from a fixed Glorot initialization~\cite{Glorot2010_Understandingdifficultytraining}. The training time of Picard-KKT-hPINN on GPU hardware is comparable to that of MLP and KKT-hPINN, averaging only 15\% longer, primarily due to the matrix inversion operation. At inference, the computational time of the three models is equivalent, even on CPU hardware (11th Gen Intel(R) Core(TM) i7-1185G7 @ 3.00GHz, 4 Core(s)).

\begin{table}[H] 
    \centering
    \caption{Accuracy comparison over the three different models. We train all the models on a GPU hardware (NVIDIA GeForce RTX 3090).}
    \begin{tabular}{lccc} 
        \hline
        & MLP & KKT-hPINN & Picard-KKT-hPINN \\ \hline
        $R^2$ [-] & 0.97 & 0.98 & 0.98 \\
        MAPE [\%] & 0.81 & 0.58 & 0.71 \\ \hline
    \end{tabular}
    \label{tab:Accuracy}
\end{table}

The accuracy of the models on the test set is comparable in terms of coefficient of determination $R^2$ and mean absolute percentage error (MAPE). In Table~\ref{tab:Accuracy}, the metrics are averaged over all the predicted variables in the test set. We do not expect the constraints to increase the accuracy considerably when a traditional MLP is already very accurate.\\
The focus of the study is to demonstrate the effectiveness of Picard-KKT-hPINN in enforcing both linear and nonlinear constraints. Figure~\ref{fig:RCE} shows the relative conservation loss of the total mass and energy. The MLP, although very accurate, does not strictly satisfy the balances. As claimed by the authors (Chen, et al., 2024), KKT-hPINN effectively enforces the linear mass balances at machine-level precision (note that the plot is in logarithmic scale), however, the predictions do not strictly satisfy the nonlinear enthalpy balance. Picard-KKT-hPINN, instead, enforces mass and enthalpy balances exactly, reducing the error below 0.001\% for both.
In a real-world scenario, the amount of data is often a bottleneck when training data-hungry machine learning models such as NNs. We analyze the performance of the models in data scarcity conditions by training on uniformly sampled fractions of the original dataset.\\
We demonstrate that constrained learning can improve model performance in data scarcity conditions (Figure~\ref{fig:data-scarcity}). Specifically, enforcing atomic balance through KKT-hPINN outperforms the MLP when using between 20 and 30\% of the original training dataset. When enforcing both atomic and total enthalpy balance with Picard-KKT-hPINN, the model performance shows a positive coefficient of determination ($R^2\simeq$ 0.75)  when only 35\% of the training dataset is used, while both MLP and KKT-hPINN show a negative $R^2$ in the same data regime.

\section{Conclusion}
We propose Picard-KKT-hPINN, a method to ensure that NN predictions adhere strictly to nonlinear algebraic constraints, such as enthalpy balances. Our work extends the recent KKT-hPINN model by Chen, et al. (2024) from linear to nonlinear constraints~\cite{Chen2024_PhysicsInformedNeural}. Thereby, we address a critical limitation of surrogate models in failing to satisfy fundamental physical principles such as mass and energy conservation. 
In our proposed method, we project the NN predictions onto the feasible space defined by mass and enthalpy balances, ensuring the exact satisfaction of these laws. We tested Picard-KKT-hPINN in surrogate modeling tasks, specifically for a methanol synthesis packed bed reactor. The method guarantees atomic and enthalpy balances without adding computational overhead, while also outperforming the accuracy of standard MLPs in data-scarce regimes. These results highlight the potential of Picard-KKT-hPINN to advance the adoption of NNs in relevant chemical engineering applications, including large-scale process optimization and simulation, where developing physically consistent surrogate models is crucial for ensuring both computational efficiency and reliable outcomes.

\section*{Acknowledgements}
This research is supported by Shell Global Solutions International B.V., for which we express sincere gratitude.
\nolinenumbers
\bibliographystyle{unsrt}  
\bibliography{PicardKKThPINN}

\end{document}